\documentclass[runningheads]{llncs}
\usepackage{graphicx}
\usepackage{algorithmic}
\usepackage[ruled,vlined]{algorithm2e}
\usepackage{xcolor}
\begin{document}
\title{Cross Language Soccer Framework: An Open Source Framework for the RoboCup 2D Soccer Simulation}
\titlerunning{Cross Language Soccer Framework}
%
\author{Nader Zare\inst{1} \and
Aref Sayareh\inst{1}\inst{2} \and
Alireza Sadraii \and
Arad Firouzkouhi\inst{1}\inst{3} \and
Amilcar Soares\inst{2}}

\authorrunning{N. Zare et al.}
%
\institute{CYRUS Robotic Team, Canada \\ \and
Memorial University of Newfoundland, St. John's, Canada\\ \and
Amirkabir University of Technology, Iran\\ 
\email{nader@cyrus2d.com} \\
\email{\{asayareh, amilcarsj\}@mun.ca} \\
\email{arad.firouzkouhi@aut.ac.ir} \\ 
\url{https://www.cyrus2d.com, https://github.com/CLSFramework}
}
\maketitle              
\begin{abstract}
RoboCup Soccer Simulation 2D (SS2D) research is hampered by the complexity of existing C++-based codes like Helios, Cyrus, and Gliders, which also suffer from limited integration with modern machine learning frameworks. This development paper introduces a transformative solution: a gRPC-based, language-agnostic framework that seamlessly integrates with the high-performance Helios base code. This approach not only facilitates the use of diverse programming languages including C\#, JavaScript, and Python—but also maintains the computational efficiency critical for real-time decision-making in SS2D. By breaking down language barriers, our framework significantly enhances collaborative potential and flexibility, empowering researchers to innovate without the overhead of mastering or developing extensive base codes. We invite the global research community to leverage and contribute to the Cross Language Soccer (CLS) framework, which is openly available under the MIT License, to drive forward the capabilities of multi-agent systems in soccer simulations\footnote{https://github.com/CLSFramework}\footnote{https://github.com/CLSFramework/cross-language-soccer-framework/wiki}.

\end{abstract}

\textbf{NOTICE:} This is an accepted article to be published by Lecture Notes in Artificial Intelligence (LNCS/LNAI) series by Springer-Verlag in the proceedings of the RoboCup International Symposium.
\section{Introduction}
\begin{figure}
    \centering
    \includegraphics[width=0.7\linewidth]{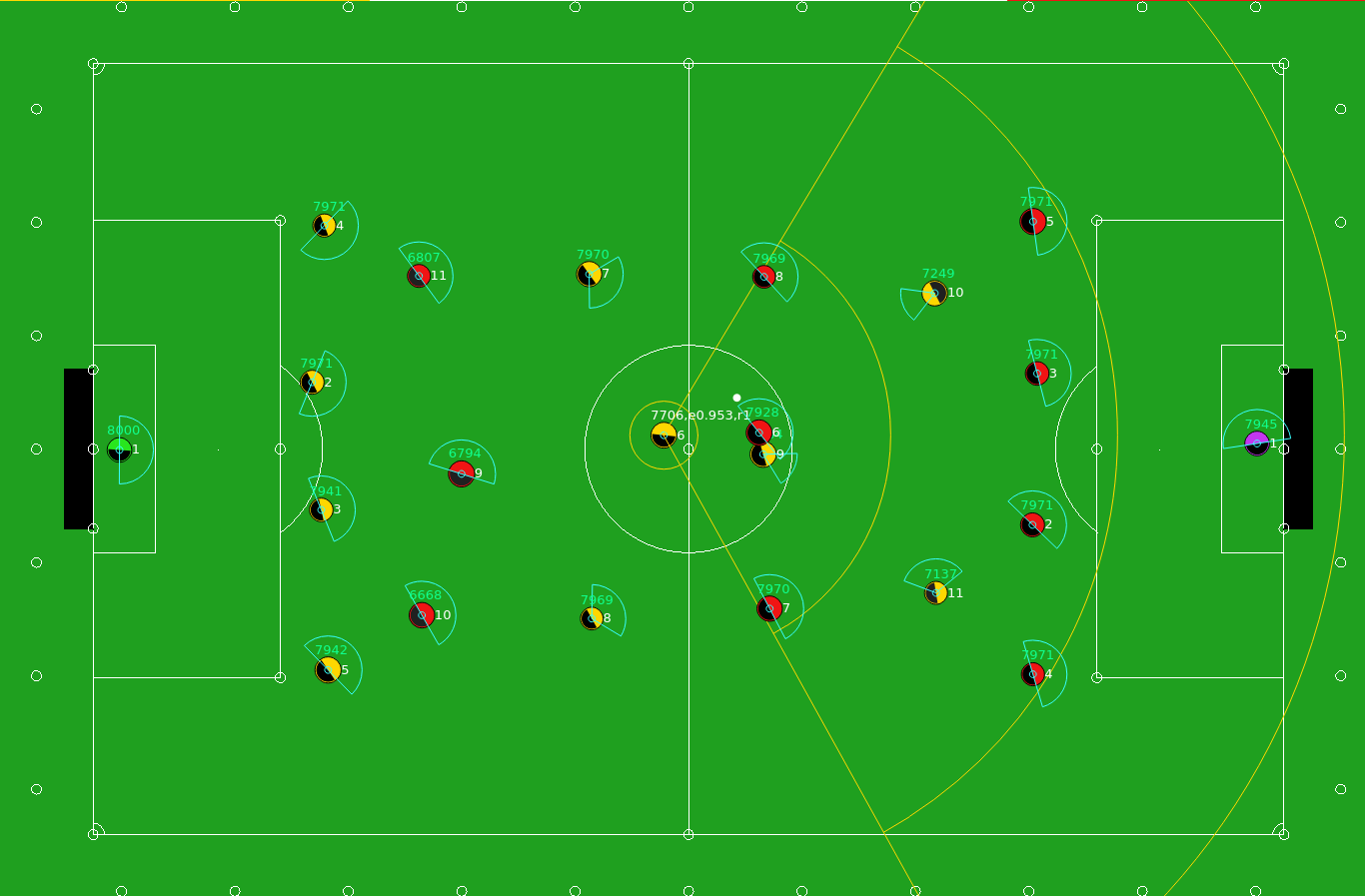}
    \caption{An overview of Soccer Simulation 2D Environment}
    \label{fig:ss2d}
\end{figure}

Soccer, a sport universally celebrated for its dynamic nature and strategic gameplay\cite{soccer1,soccer2,soccer3}, serves as a captivating domain for artificial intelligence research. 
The RoboCup organization was established to advance the abilities of autonomous agents to understand and play soccer. The ultimate objective of this organization is to create a fully autonomous team that can defeat the FIFA World Cup champion by the year 2050
\cite{noda1996soccer,burkhard2002,kitano1997robocup,robo1997}.
Soccer Simulation 2D League (SS2D) is a platform that promotes the development of artificial intelligence in multi-agent environments
\cite{ref1,ref2,ref3,ref4,ref5,ref6,ref7,ref8,cyrussamposiom}. Figure \ref{fig:ss2d} provides a detailed visualization of the Soccer Simulation 2D game environment.

Existing widely-used C++-based codes like Helios, Cyrus2D, and Gliders offer tools that allow researchers to focus on the decision challenges within the environment \cite{agent2d,cyrus2d,glbase}. Nevertheless, due to the inherent complexity of C++ and its limited integration with machine-learning frameworks, incorporating machine-learning methods into these frameworks poses significant challenges.
Addressing these challenges involves the development of new frameworks in alternative languages. However, creating frameworks in different languages is a time-consuming process, and the performance of these languages may not be sufficiently high for making timely decisions. For example, Python, a widely used and straightforward programming language, lacks the necessary performance capabilities in the context of SS2D\cite{pyrus}.

Recognizing the advantages of a high-performance base code, such as Helios, which completely supports SS2D server features, we propose a novel solution. This development paper introduces an innovative approach to overcome language barriers and enhance flexibility. By connecting the robust Helios base code to other languages using gRPC, we aim to provide researchers with the freedom to work in their preferred languages without compromising the performance benefits offered by C++. 
This approach empowers researchers to focus on the intelligence aspects of their agents, leveraging the high-quality and high-performance foundation laid by Helios.

As illustrated in Figure \ref{fig:enter-label}, the Cross Language Soccer framework architecture seamlessly integrates the Helios base code with a language-agnostic gRPC server. This connection facilitates efficient communication and collaboration between the high-performance Helios base code and multiple programming languages, empowering researchers in the RoboCup 2D Soccer Simulation domain.

\begin{figure}
    \centering
    \includegraphics[width=0.7\linewidth]{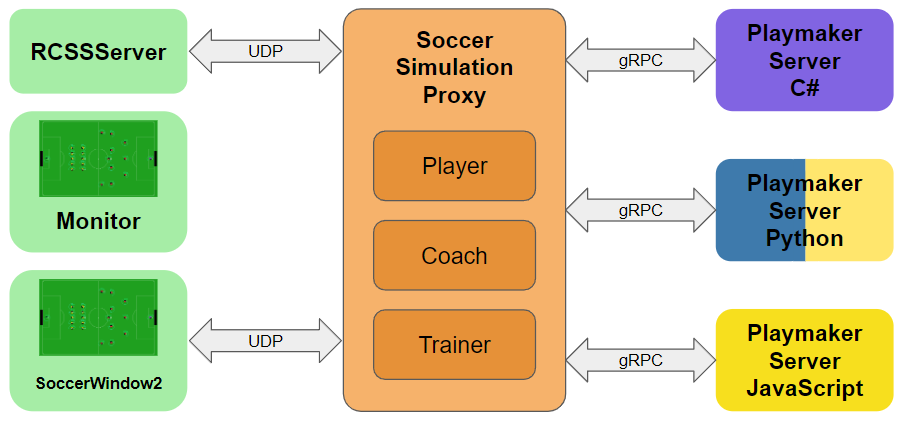}
    \caption{CLS Framework Architecture: The diagram depicts the interconnected components of the CLS framework approach. The modified version of the Helios base code called Soccer Simulation Proxy communicates with a language-agnostic gRPC Playmaker Server, providing a flexible environment for researchers working with various programming languages in RoboCup 2D Soccer Simulation.}
    \label{fig:enter-label}
\end{figure}

In the following sections, 
we provide a detailed explanation of our approach, including the framework's architecture, implementation details, and its implications for research in the SS2D environment.

\section{Background}
The RoboCup Soccer Simulation 2D server (server) has two main jobs: it runs soccer games and interacts with coaches, players, trainers, and monitors.
It communicates by sending observations to agents, receiving low-level actions from them, and applying them to the game.
Various teams have contributed base codes to the RoboCup soccer simulation, for instance, the "CMUnited" team from Carnegie Mellon University (USA)\cite{stonerobo2000,stonerobo99}, the "UvA Trilearn" team from the University of Amsterdam (The Netherlands)\cite{UTR}, the "MarliK" team from the University of Guilan (Iran)\cite{MKR}, the "HELIOS" team from AIST Information Technology Research Institute (Japan), the "Wrighteagle" team from the University of Science and Technology of China\cite{wrighteagle}, the "Gliders2d" team from the University of Sydney (Australia)\cite{glbase,glbase2}, and the "Cyrus2d" and "Pyrus" teams from Dalhousie University (Canada)\cite{cyrussamposiom,cyrus2d,pyrus}.

The Helios base code provides in processing observations by denoising them and constructing a detailed world model that contains all of the information required for decision-making, such as an intercept table.
Also, its richness in functionality, encompasses high-level actions (e.g., BodyGoToPoint, BodySmartKick), offensive planners (e.g. ChainActions), and extensive logging functionality, offering graphical logs through the SoccerWindow2 application.

In navigating the complexities of existing base codes like Cyrus and Gliders, which are built upon the powerful Helios base code, researchers often face challenges due to their intricacy. While these codes provide robust capabilities, applying new algorithms or working with them requires a deep understanding of both soccer simulation 2D and the base codes. Moreover, their reliance on C++ poses hurdles, as the language lacks abundant machine-learning frameworks and is relatively challenging for researchers less familiar with SS2D or C++. On the other hand, the Pyrus base code, developed in Python, offers a more accessible option but sacrifices performance.

Half Field Offense (HFO) is an environment tailored for reinforcement learning in soccer simulations, focusing exclusively on half-field scenarios\cite{hfo}. This framework modifies the standard Soccer Simulation 2D server, limiting it to half-field play, which restricts its applicability to full-match scenarios and broader training objectives. Furthermore, HFO only supports C++ and Python, excluding other programming languages. Its specialized nature and lack of support for recent server updates or features also restrict its utility for researchers looking to utilize the latest developments in soccer simulation.

To address these challenges, we propose a novel solution by using Remote Procedure Call (RPC) technology, specifically gRPC (Google Remote Procedure Call). We implemented the gRPC client on the Helios base, so it facilitates efficient communication between the base and other languages.
Therefore, we streamline the development process and enhance the flexibility of researchers. This approach allows developers to benefit from the high performance of C++ for computationally intensive tasks, such as denoising and world modeling, while enabling them to implement or improve Helios base decision-making logic in languages like C\#, JavaScript, Python, or others.

The use of gRPC offers several advantages, including low latency, language-agnostic communication, and efficient serialization of data. This integration helps researchers work in their preferred languages without sacrificing the performance advantages of the Helios base code. In summary, our solution facilitates a bridge between the efficiency of C++ and the flexibility of other languages, empowering researchers in the SS2D domain.

\section{Cross Language Soccer Framework}
We have introduced a new approach called Cross Language Soccer Framework to enhance the flexibility and interoperability of RoboCup Soccer Simulation 2D (SS2D). The Cross Language Soccer approach consists of two components: Soccer Simulation Proxy and Playmaker Server.
The Soccer Simulation Proxy is an extended version of the Helios base that can send decision-making information to the Playmaker Server. It can receive high-level actions from the Playmaker Server and send them to RoboCup Soccer Simulation Server and/or SoccerWindow2.
On the other hand, the Playmaker Server receives information from the client and selects the appropriate actions to be sent back to the client. We have implemented some sample servers in C\#, Python, and JavaScript, but it can also be implemented in other languages to make use of their features.

\subsection{Soccer Simulation Proxy}
We have modified the Helios base code to enhance its interaction capabilities with the gRPC server\footnote{https://github.com/CLSFramework/soccer-simulation-proxy}. Specifically, the base code now transmits detailed information about each game cycle to the server, which in response, sends back a series of potential actions. This bidirectional communication enables the Playmaker Server to effectively process and make strategic decisions based on the incoming data.

To facilitate this enhanced functionality, the client is designed to serialize various data types—such as the WorldModel, FullState WorldModel, Server Parameters, Player Parameters, and Player Types—into gRPC messages during each game cycle. Conversely, it also deserializes incoming gRPC action messages back into Helios base actions.

These improvements ensure that the Helios base maintains its inherent efficiency and performance, while now being able to seamlessly interact with a diverse array of programming languages through gRPC. This modification not only preserves the robustness of the original system but also expands its utility and applicability in multi-language research environments.

\subsection{Playmaker Server}
The Playmaker Server acts as the decision-making hub within our framework, receiving and processing messages from the Soccer Simulation Proxy. Based on the detailed information conveyed in these messages, it evaluates the current game state and strategically generates corresponding actions to be executed in the simulation.

To demonstrate the versatility and language-agnostic design of our framework, we have developed initial implementations of the Playmaker Server in three different programming languages: C\#\footnote{https://github.com/CLSFramework/playmaker-server-csharp}, Python\footnote{https://github.com/CLSFramework/playmaker-server-python}, and JavaScript\footnote{https://github.com/CLSFramework/playmaker-server-nodejs}. These examples serve to showcase the server’s capability to operate effectively across diverse programming environments. Looking ahead, we aim to extend this functionality by developing additional server implementations in other languages, further broadening the framework's accessibility and appeal to a global research community.

\subsection{Benefits of using gRPC}
To facilitate robust communication between the Playmaker Server and the Soccer Simulation Proxy, we employ gRPC, a high-performance, open-source universal RPC framework. The core of this communication setup is a Protocol Buffer, which defines the structure of messages and services. This Protocol Buffer is instrumental in generating an abstract server and client-side objects for each message, utilizing libraries that support multiple programming languages.

The adoption of gRPC streamlines the integration process for researchers working in different programming environments. It eliminates the cumbersome need for manual serialization and deserialization of messages. Researchers can directly utilize the automatically created objects for decision-making and action generation, which are then seamlessly communicated to the SS2D-Proxy-Client. This efficient data handling mechanism not only accelerates development but also enhances the flexibility and scalability of server implementations across diverse research contexts.

\subsection{Protocol Buffer Definitions}
The communication protocol between the modified version of the Helios base and the Playmaker Server is defined using Protocol Buffers (ProtoBuf). The gRPC service 'Game' comprises several Remote Procedure Call (RPC) functions for obtaining player, coach, and trainer actions, as well as sending initialization messages and server parameters. The 'State' message encapsulates crucial information about the game state, including agent type and a detailed world model. Figure \ref{fig:enter-label3} provides an overview of the 'Game' service and highlights key messages utilized in the Protocol Buffer. Detailed explanations of the RPC functions and messages are presented in the subsequent sections.
\begin{figure}
    \centering
    \includegraphics[width=0.7\linewidth]{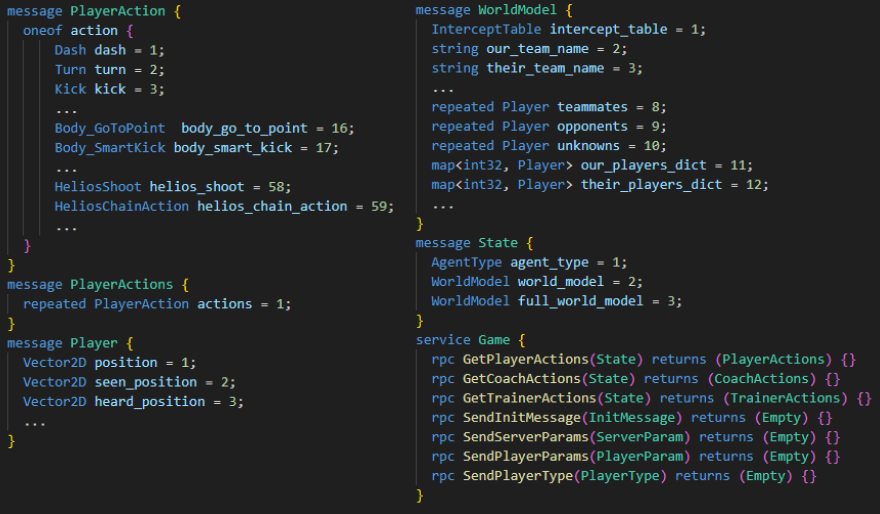}
    \caption{Definition of Protocol Buffer Messages and RPC Functions for the 'Game' Service in SS2D-Proxy. The illustration outlines essential structures and functions used for communication between the Helios base and the SS2D-Proxy-Server.}
    \label{fig:enter-label}
\end{figure}

\subsection{Key RPC Functions}
\begin{itemize}
\item \textbf{GetPlayerActions(State)}: Returns a list of player actions based on the provided game state.
\item \textbf{GetCoachActions(State)}: Retrieves coach actions corresponding to the given game state.
\item \textbf{GetTrainerActions(State)}: Fetches trainer actions based on the provided game state.
\item \textbf{SendInitMessage(InitMessage)}: Sends an initialization message and receives an acknowledgment.
\item \textbf{SendServerParams(ServerParam)}: Transmits server parameters and acknowledges receipt.
\item \textbf{SendPlayerParams(PlayerParam)}: Sends player parameters and receives an acknowledgment.
\item \textbf{SendPlayerType(PlayerType)}: Sends player type information and acknowledges receipt.
\end{itemize}

\subsection{Message Definitions}
\begin{itemize}
    \item \textbf{State}: The 'State' message encapsulates the agent type and a detailed 'WorldModel,' containing essential game state information such as team names, player details, ball position, game mode, and scores\footnote{https://github.com/CLSFramework/cross-language-soccer-framework/wiki/Environment-Messages}.
    \item \textbf{PlayerAction}: The 'PlayerActions' message consists of a list of 'PlayerAction' messages, each representing an action that a player can perform during a game cycle. These actions cover a wide range of behaviours, including movement, ball interaction, communication, and specific strategies like those implemented in the Helios base code.
\item \textbf{TrainerAction}: Represents actions initiated by a trainer during the game.
\item \textbf{CoachAction}: Represents actions initiated by a coach during the game.
\item \textbf{ServerParam}: Contains parameters related to the game server, contributing to the game's configuration.
\item \textbf{PlayerParam}: Encompasses parameters specific to individual players, influencing their behaviour in the game.
\item \textbf{PlayerType}: Provides insight into a player's style, helping to differentiate and classify players on the field.
\end{itemize}

\begin{figure}[h] 
    \centering
    \includegraphics[width=1\linewidth]{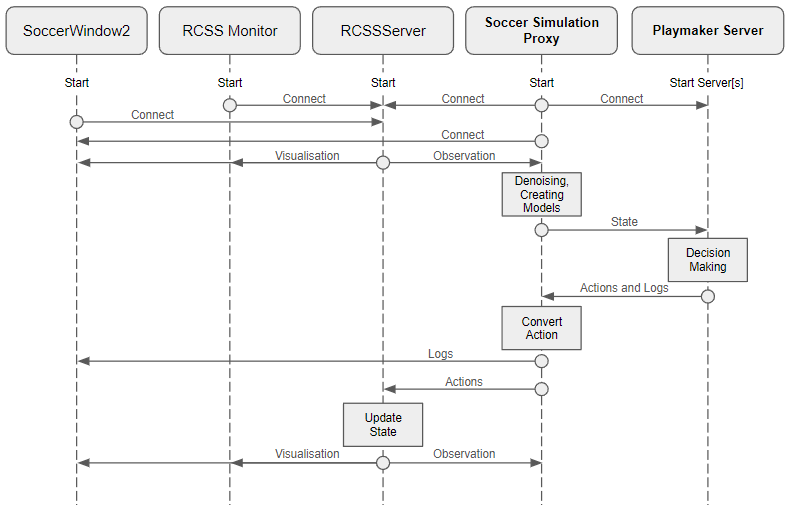}
    \caption{SS2D-Proxy Message Flow: The diagram illustrates the message flow between key components, showcasing the interconnected communication pathways among the SS2D Proxy Base, SS2D Proxy Server, Soccer Window 2 Monitor, and the SS2D Server in the development of the SS2D-Proxy framework.}
    \label{fig:enter-label2}
\end{figure}

As depicted in Figure \ref{fig:enter-label2}, the message flow illustrates the seamless communication between the SS2D Proxy Base, SS2D Proxy Server, Soccer Window 2 Monitor, and the SS2D Server. This visual representation offers insights into the data exchange dynamics within the proposed SS2D-Proxy framework.

In summary, the SS2D-Proxy approach fosters collaboration and flexibility in RoboCup SS2D research, enabling seamless communication between the high-performance Helios base code and diverse programming languages through the gRPC protocol. This integration lays the groundwork for advancing intelligent multi-agent systems in the dynamic domain of RoboCup Soccer Simulation 2D.

\section{Development Features}

This section highlights key development challenges encountered while building the framework and the innovative solutions implemented to address these challenges. These solutions are designed to simplify the setup and use of our framework, enhancing its accessibility for researchers and participants in the RoboCup competition.

\subsection{Cross-Platform Compatibility Challenges}

\textbf{Challenge:} Setting up the Soccer Simulation Server, Soccer Simulation Proxy, and Soccer Simulation Monitor on Ubuntu was initially complex due to multiple dependencies and configuration requirements.

\textbf{Solution:} We provide Linux AppImages for the Soccer Simulation Server, Soccer Simulation Proxy, and Monitor. These AppImages are self-contained and include all necessary components and dependencies, allowing users to download and execute them with ease. To customize or extend the functionality of the Playmaker Server, researchers simply need to install language-specific dependencies using NPM for JavaScript, PIP for Python, or .NET for C\#. This approach significantly simplifies the initial setup process, enabling researchers to quickly start developing and testing their strategies.

\subsection{Windows Compatibility}

\textbf{Challenge:} Building and running the Soccer Simulation Server, Soccer Simulation Proxy, and Monitor on Windows platforms is not directly supported due to compatibility issues.

\textbf{Solution:} We recommend using Windows Subsystem for Linux (WSL) to run the server components, while the Playmaker Server can run natively on Windows. Alternatively, we provide a Dockerized version of the framework, allowing users to deploy the Soccer Simulation components in Docker containers and connect them seamlessly to the Playmaker Server running on Windows.

\subsection{RoboCup Competition Preparation}

\textbf{Challenge:} Participants in the RoboCup competition need a reliable and easy-to-setup environment to compete without technical issues.

\textbf{Solution:} We offer an AppImage of the Soccer Simulation Proxy that integrates smoothly into the Playmaker project. This setup ensures that teams can participate in the RoboCup competition with minimal setup and without compatibility concerns.

\subsection{Support for Reinforcement Learning Research}

\textbf{Challenge:} Researchers aiming to implement reinforcement learning algorithms require a flexible and robust environment compatible with modern RL frameworks.

\textbf{Solution:} Our framework includes sample Python code that researchers can use as a foundation for training agents. Additionally, we provide mechanisms for controlling the Soccer Simulation Server via a trainer agent, facilitating the implementation and testing of advanced RL algorithms.

\subsection{Integration with Gym Environments}

\textbf{Challenge:} Researchers familiar with developing RL algorithms in gym environments need a comparable setup to apply their expertise in the soccer simulation context\cite{gym}.

\textbf{Solution:} We have developed sample projects that mimic the design of gym environments, allowing researchers to leverage existing RL libraries and techniques. This approach provides a familiar and efficient starting point for developing new RL strategies in our soccer simulation framework.

\section{Conclusion}
In conclusion, the SS2D-Proxy approach presented in this paper addresses the challenges associated with existing C++-based codes in RoboCup Soccer Simulation 2D (SS2D) research. The complexities and learning difficulties posed by these codes, coupled with the time-consuming process of developing new base codes for each language, motivated the need for a versatile solution.

Our innovation involves connecting the widely adopted Helios base code to a language-agnostic framework using gRPC. This approach facilitates seamless communication between Helios and different languages, such as C\#, Python, and JavaScript. By doing so, we empower researchers to work in their language of choice without compromising the high-quality, high-performance features offered by the C++ Helios base code.

The integration of gRPC not only streamlines communication but also introduces efficiency through low-latency data exchange and language-agnostic capabilities. Researchers can now focus on developing intelligent multi-agent systems in SS2D without the burdensome task of creating an entire base code.

Our solution enhances collaboration, allowing researchers to leverage the strengths of C++ for computationally intensive tasks, while developing decision-making logic in languages that better suit their preferences and expertise. The protocol buffer definitions and RPC functions provide a standardized and efficient means of communication between Helios and the gRPC server, ensuring a seamless exchange of game state information and actions.

In summary, the SS2D-Proxy approach opens new avenues for collaborative research in the RoboCup SS2D domain. It bridges the gap between the efficiency of C++ and the flexibility of multiple languages, paving the way for advancements in intelligent multi-agent systems within the dynamic context of RoboCup Soccer Simulation 2D.

\section*{Acknowledgments}
We extend our heartfelt appreciation to Professor Stan Matwin and Professor Amilcar Soares for their exceptional support and guidance over the past three years. Their mentorship has been instrumental in shaping the trajectory of our research in the CYRUS team.


Special thanks are extended to Professor Hidehisa Akyama for his pivotal work on the Helios base and Soccer Simulation 2D server. His pioneering efforts have laid a solid foundation for researchers in the field, including ourselves.

These acknowledgments stand as a tribute to the collaborative spirit and mentorship that have played a crucial role in the development and realization of this project.


\begin{thebibliography}{8}
\bibitem{soccer1}
Bangsbo J, Peitersen B. Soccer systems and strategies. Human Kinetics; 2000.

\bibitem{soccer2}
Pollard R, Reep C. Measuring the effectiveness of playing strategies at soccer. Journal of the Royal Statistical Society: Series D (The Statistician). 1997 Dec;46(4):541-50.

\bibitem{soccer3}
Rein R, Memmert D. Big data and tactical analysis in elite soccer: future challenges and opportunities for sports science. SpringerPlus. 2016 Dec;5(1):1-3.

\bibitem{burkhard2002}
Burkhard, H.D., Duhaut, D., Fujita, M., Lima, P., Murphy, R., Rojas, R.: The road to RoboCup 2050. IEEE Robotics Automation Magazine 9(2), 31–38 (Jun 2002)


\bibitem{noda1996soccer}
Noda, I. and Matsubara, H., 1996, November. Soccer server and researches on multi-agent systems. In Proceedings of the IROS-96 Workshop on RoboCup (pp. 1-7).

\bibitem{robo1997}
Kitano, H., Asada, M., Kuniyoshi, Y., Noda, I. and Osawa, E., 1997, February. Robocup: The robot world cup initiative. In Proceedings of the first international conference on Autonomous agents (pp. 340-347).

\bibitem{kitano1997robocup}
Kitano, H., Asada, M., Kuniyoshi, Y., Noda, I., Osawa, E. and Matsubara, H., 1997. RoboCup: A challenge problem for AI. AI magazine, 18(1), pp.73-73.

\bibitem{ref1}
Noda, I., Stone, P.: The RoboCup Soccer Server and CMUnited Clients: Implemented Infrastructure for MAS Research. Autonomous Agents and Multi-Agent Systems 7(1–2), 101–120 (July–September 2003)

\bibitem{ref2}
Riley, P., Stone, P., Veloso, M.: Layered disclosure: Revealing agents’ internals. In: Castelfranchi, C., Lesperance, Y. (eds.) Intelligent Agents VII. Agent Theories, Architectures, and Languages — 7th. International Workshop, ATAL-2000, Boston, MA, USA, July 7–9, 2000, Proceedings. Lecture Notes in Artificial Intelligence, Springer, Berlin, Berlin (2001)

\bibitem{ref3}
Stone, P., Riley, P., Veloso, M.: Defining and using ideal teammate and opponent models. In: Proc. of the 12th Annual Conf. on Innovative Applications of Artificial Intelligence (2000)

\bibitem{ref4}
Butler, M., Prokopenko, M., Howard, T.: Flexible synchronisation within RoboCup environment: A comparative analysis. In: RoboCup 2000: Robot Soccer World Cup IV. pp. 119–128. Springer, London, UK (2001)

\bibitem{ref5}
Reis, L.P., Lau, N., Oliveira, E.: Situation based strategic positioning for coordinating a team of homogeneous agents. In: Balancing Reactivity and Social Deliberation in Multi-Agent Systems, From RoboCup to Real-World Applications. pp. 175–197. Springer (2001)

\bibitem{ref6}
Prokopenko, M., Wang, P.: Disruptive Innovations in RoboCup 2D Soccer Simulation League: From Cyberoos’98 to Gliders2016. In: Behnke, S., Sheh, R., Sariel, S., Lee, D.D. (eds.) RoboCup 2016: Robot World Cup XX [Leipzig, Germany, June 30 - July 4, 2016]. Lecture Notes in Computer Science, vol. 9776, pp. 529–541. Springer (2017)

\bibitem{ref7}
Prokopenko, M., Wang, P., Marian, S., Bai, A., Li, X., Chen, X.: Robocup 2d soccer simulation league: Evaluation challenges. In: Akiyama, H., Obst, O., Sammut, C., Tonidandel, F. (eds.) RoboCup 2017: Robot World Cup XXI [Nagoya, Japan, July 27-31, 2017]. Lecture Notes in Computer Science, vol. 11175, pp. 325–337. Springer (2018) 

\bibitem{ref8}
Prokopenko, M., Wang, P., Obst, O.: Gliders2015: Opponent avoidance with bio-inspired flocking behaviour. In: RoboCup 2015 Symposium and Competitions: Team Description Papers, Hefei, China, July 2015 (2015)

\bibitem{cyrussamposiom}
Zare, N., Sayareh, A., Sarvmaili, M., Amini, O., Matwin, S., Soares, A.: Engineering Features to Improve Pass Prediction in 2D Soccer Simulation Games. In: RoboCup 2021: Robot World Cup XXIV, Springer (2021)

\bibitem{agent2d}
Akiyama, H., Nakashima, T.: Helios base: An open source package for the robocup soccer 2d simulation. In Robot Soccer World Cup 2013 Jun 24 (pp. 528-535). Springer, Berlin, Heidelberg.

\bibitem{cyrus2d}
Zare, N., Amini, O., Sayareh, A., Sarvmaili, M., Firouzkouhi, A., Rad, S.R., Matwin, S. and Soares, A., 2023. Cyrus2D Base: Source Code Base for RoboCup 2D Soccer Simulation League. In RoboCup 2022: Robot World Cup XXV (pp. 140-151). Cham: Springer International Publishing.

\bibitem{stonerobo2000}
Stone, P., Asada, M., Balch, T.R., Fujita, M., Kraetzschmar, G.K., Lund, H.H., Scerri, P., Tadokoro, S., Wyeth, G.: Overview of robocup-2000. In: Stone, P., Balch, T.R., Kraetzschmar, G.K. (eds.) RoboCup 2000: Robot Soccer World Cup IV. Lecture Notes in Computer Science, vol. 2019, pp. 1–28. Springer (2000)

\bibitem{stonerobo99}
Stone, P., Riley, P., Veloso, M.: The CMUnited-99 champion simulator team. In: Veloso, M., Pagello, E., Kitano, H. (eds.) RoboCup-99: Robot Soccer World Cup III, Lecture Notes in Artificial Intelligence, vol. 1856, pp. 35–48. Springer Verlag, Berlin (2000)

\bibitem{UTR}
Kok, J.R., Vlassis, N., Groen, F.: UvA Trilearn 2003 team description. In: Polani, D., Browning, B., Bonarini, A., Yoshida, K. (eds.) Proceedings CD RoboCup 2003. Springer, Padua (2003)

\bibitem{wrighteagle}
Bai A, Chen X, MacAlpine P, Urieli D, Barrett S, Stone P. Wrighteagle and ut austin villa: RoboCup 2011 simulation league champions. InRoboCup 2011: Robot Soccer World Cup XV 15 2012 (pp. 1-12). Springer Berlin Heidelberg.

\bibitem{glbase}
Prokopenko, M., Wang, P.: Gliders2d: Source Code Base for RoboCup 2D Soccer Simulation League. CoRR abs/1812.10202 (2018)

\bibitem{glbase2}
Prokopenko, M. and Wang, P., 2019, July. Fractals2019: Combinatorial optimisation with dynamic constraint annealing. In Robot World Cup (pp. 616-630). Springer, Cham.

\bibitem{MKR}
Tavafi, A., Nozari, N., Vatani, R., Yousefi, M.R., Rahmatinia, S., Pirdir, P.: MarliK 2012 Soccer 2D Simulation Team Description Paper. In: RoboCup 2012 Symposium and Competitions: Team Description Papers, Mexico City, Mexico, June 2012 (2012) 

\bibitem{ggsoccer}
Kurach, K., Raichuk, A., Sta\'nczyk, P., Zajac, M., Bachem, O., Espeholt, L., Riquelme, C., Vincent, D., Michalski, M., Bousquet, O. and Gelly, S., 2020, April. Google research football: A novel reinforcement learning environment. In Proceedings of the AAAI Conference on Artificial Intelligence (Vol. 34, No. 04, pp. 4501-4510).


\bibitem{hfo}
Hausknecht, M., Mupparaju, P., Subramanian, S., Kalyanakrishnan, S. and Stone, P., 2016, May. Half field offense: An environment for multiagent learning and ad hoc teamwork. In AAMAS Adaptive Learning Agents (ALA) Workshop (Vol. 3). sn.

\bibitem{gym}
Brockman, G., Cheung, V., Pettersson, L., Schneider, J., Schulman, J., Tang, J. and Zaremba, W., 2016. Openai gym. arXiv preprint arXiv:1606.01540.

\bibitem{pyrus}
Zare, N., Sayareh, A., Amini, O., Sarvmaili, M., Firouzkouhi, A., Matwin, S. and Soares, A., 2023. Pyrus Base: An Open Source Python Framework for the RoboCup 2D Soccer Simulation. arXiv preprint arXiv:2307.16875.

\end{thebibliography}
\end{document}